\documentclass[conference]{IEEEtran}

\IEEEoverridecommandlockouts
\usepackage{cite}
\usepackage{amsmath,amssymb,amsfonts}
\usepackage{graphicx}
\usepackage{textcomp}
\usepackage{xcolor}
\usepackage{algorithm}
\usepackage{algpseudocode}
\usepackage{bbm}

\usepackage{array}
\usepackage{tabularx}
\usepackage{float}
\IEEEoverridecommandlockouts

\usepackage{tikz}
\usetikzlibrary{svg.path}
\usepackage[hidelinks]{hyperref}
\usepackage{scalerel}
\usepackage{enumitem}
\definecolor{orcidlogocol}{HTML}{A6CE39}
\tikzset{
  orcidlogo/.pic={
    \fill[orcidlogocol] svg{M256,128c0,70.7-57.3,128-128,128C57.3,256,0,198.7,0,128C0,57.3,57.3,0,128,0C198.7,0,256,57.3,256,128z};
    \fill[white] svg{M86.3,186.2H70.9V79.1h15.4v48.4V186.2z}
                 svg{M108.9,79.1h41.6c39.6,0,57,28.3,57,53.6c0,27.5-21.5,53.6-56.8,53.6h-41.8V79.1z M124.3,172.4h24.5c34.9,0,42.9-26.5,42.9-39.7c0-21.5-13.7-39.7-43.7-39.7h-23.7V172.4z}
                 svg{M88.7,56.8c0,5.5-4.5,10.1-10.1,10.1c-5.6,0-10.1-4.6-10.1-10.1c0-5.6,4.5-10.1,10.1-10.1C84.2,46.7,88.7,51.3,88.7,56.8z};
  }
}

\newcommand\orcidicon[1]{\href{https://orcid.org/#1}{\mbox{\scalerel*{
\begin{tikzpicture}[yscale=-1,transform shape]
\pic{orcidlogo};
\end{tikzpicture}
}{|}}}}

\usepackage{amsmath,amssymb}
\usepackage{booktabs}
\usepackage{graphicx}
\usepackage{multirow}
\usepackage{url}
\usepackage{xcolor}

\newcommand{\framework}{\textsc{TwinGridShield}}
\newcommand{\safe}{\mathcal{S}}

\begin{document}
\title{TwinGridShield: Consequence-Aware Runtime Authorization for LLM Grid-Agent Actions}


\author{\IEEEauthorblockN{Md~Fazley~Rafy$^{\textsuperscript{\orcidicon{0000-0003-3057-9546}}}$}
\IEEEauthorblockA{\textit{Lane Department of Computer Science and Electrical Engineering} \\
\textit{West Virginia University}\\
Morgantown, WV-26505, USA \\
ORCID: 0000-0003-3057-9546}
\thanks{Author is with the Lane Department of Computer Science and Electrical Engineering, West Virginia University, Morgantown, WV, USA, 26505. Corresponding Author email address: mdfazleyrafy@ieee.org.}}

\maketitle

\IEEEpubid{\makebox[\columnwidth]{979-8-3315-5720-1/26/\$31.00 ©2026 IEEE\hfill}\hspace{\columnsep}\makebox[\columnwidth]{}} 
\IEEEpubidadjcol

\begin{abstract}
Large language model (LLM)-assisted energy-management tools can translate natural-language context into structured grid commands, but syntactic validity does not imply physical admissibility. This paper presents TwinGridShield, a model-independent runtime authorization layer that evaluates each proposed action in a deterministic network twin before release. The prototype checks connectivity, branch-flow, generator, and load-shedding invariants and records each decision in a hash-chained log. A controlled IEEE 14-bus study evaluates single-step switching, redispatch, and load-shedding actions using DC power flow and experimentally assigned branch ratings. In the matched-model experiment, a stochastic proposal source configured to select an unsafe action with probability $p=0.84$ produced 421 unsafe proposals in 500 attacked-condition trials, a realized rate of 84.2\%. This value characterizes the configured surrogate and is not an empirical measurement of LLM prompt-injection susceptibility. TwinGridShield produced 0 unsafe releases in those 500 trials. Because action labeling and authorization used the same DC model, system state, branch ratings, and encoded constraints, this result verifies conformance of the implementation to its encoded authorization predicate rather than safety under model error. The principal robustness evaluation therefore introduces model mismatch. Unsafe acceptance reached 5.63\% under bounded $\pm20\%$ per-bus load-measurement error and 30.09\% when actual branch ratings were 20\% below modeled ratings.
\end{abstract}


\section{Introduction}
\IEEEPARstart{L}{LM}-based assistants are moving from explaining operating procedures toward calling software tools. In a power-system domain, an agent might query an energy management system (EMS), summarize alarms, formulate a redispatch, or prepare a switching request. Domain models already demonstrate that language models can support advanced power-dispatch tasks~\cite{cheng2025gaia}, and the reasoning-and-action pattern explicitly interleaves language reasoning with external tool calls~\cite{yao2023react}. This interface can map heterogeneous data and operator intent into a common conversational workflow, but it also changes the failure boundary. A wrong answer is no longer only text; it can become an action against cyber-physical infrastructure. Two weaknesses meet at that boundary. First, token prediction is not a certified model of network physics. An agent can output a correctly typed branch-opening, redispatch, or load-shedding command while missing islanding, thermal-limit, generator-limit, or critical-service consequences, which is often referred to as action blindness~\cite{wang-etal-2026-action}. As a concrete example, a tampered work order may instruct an agent to open a heavily loaded branch to clear a diagnostic code. The resulting command can satisfy the tool schema while islanding part of the network or transferring flow beyond a branch rating, without triggering any schema-level alarm. Second, the context presented to an agent mixes trusted instructions with untrusted content. Indirect prompt injection has been shown to compromise LLM-integrated applications through retrieved or observed text~\cite{greshake2023indirect}. Direct instruction hierarchies can also be subverted by crafted prompts~\cite{perez2022ignore}. In a control center, the untrusted channel can be an equipment description, vendor document, ticket, alarm annotation, or compromised retrieval record. Network segmentation and role-based access remain necessary, but access authorization alone does not answer the state-dependent physical question: is this particular command safe under the current grid condition? 

Power-grid security is already a cyber-physical problem rather than a purely information-security problem~\cite{sridhar2012cyber}. Industrial-control guidance emphasizes defense in depth and consequence-aware risk management~\cite{nist80082}, while the NIST generative-AI profile calls for measuring and managing risks across the AI lifecycle~\cite{nistai6001}. Existing LLM defenses often focus on classifying malicious text or constraining tool syntax. Text filters are sensitive to phrasing, while a schema establishes only that a command is well formed. Agent sandboxes can expose risky behavior before real execution~\cite{ruan2023toolemu}, but an LM-emulated environment is not a deterministic power-flow certificate. Conversely, classical runtime enforcement can suppress events that violate a security policy~\cite{schneider2000enforce}, and shielding has been used to constrain learned control policies~\cite{alshiekh2018shielding}. The remaining need is a compact architecture that applies runtime-enforcement principles to tool-using grid agents while remaining independent of the agent's prompt, vendor, and model.
The contribution of this paper is three-fold:
\begin{enumerate}
\item Formalizes action blindness for a grid agent as a mismatch between a proposed structured action and a consequence-closed safe set.
\item Develops \framework{}, which combines typed commands, a deterministic network twin, explicit operating invariants, fail-closed enforcement, escalation, and a tamper-evident decision chain.
\item Evaluates \framework{} in a controlled IEEE 14-bus authorization study using a configured stochastic proposal source, matched-model conformance testing, runtime measurements, and a separate parameter-mismatch analysis where the surrogate experiment evaluates authorization policies under a fixed proposal distribution
\end{enumerate}

\section{Related Work and Research Gap}

\begin{table}[t]
\caption{Representative LLM agent systems in power-system operations}
\label{tab:llm-survey}
\centering
\footnotesize
\setlength{\tabcolsep}{4pt}
\begin{tabular}{llcc}
\toprule
System & Primary task & Action- & Runtime \\
&              & executing? & safety gate? \\
\midrule
GAIA~\cite{cheng2025gaia}             & Dispatch optimization   & Yes & No explicit \\
PowerAgent~\cite{zhang2025poweragent} & Agentic grid workflow   & Yes & No explicit \\
Grid-Agent~\cite{zhu2025gridagent}    & Violation remediation   & Yes & LM sandbox \\
GridMind~\cite{jin2025gridmind}       & System analysis         & Yes & No explicit \\
ChatGrid~\cite{jin2024chatgrid}       & Visualization / Q\&A    & No  & N/A \\
\textbf{This work}                    & Action authorization    & --- & \textbf{Prop.~1} \\
\bottomrule
\end{tabular}
\end{table}

As Table~\ref{tab:llm-survey} shows, LLM agents in power systems have moved from advisory Q\&A toward tool-mediated operation. Here, no explicit indicates that the cited work does not report a deterministic runtime gate that checks the physical consequence of each proposed grid action before release. Tool-using frameworks such as ReAct~\cite{yao2023react} interleave natural-language reasoning with structured external calls, and domain agents now span dispatch optimization, violation remediation, and multi-fidelity simulation support~\cite{cheng2025gaia,zhang2025poweragent,zhu2025gridagent}. The safety boundary changes when workflow automation converts a language-model recommendation into a machine-authorized command, or when repeated suggestions create automation bias. The relevant assurance problem then includes the model, context sources, tool broker, network model, actuator permissions, and operator, rather than model accuracy alone.

Perez and Ribeiro analyze direct instruction attacks against language models~\cite{perez2022ignore}. Greshake {et al.} examine the operationally relevant indirect case, in which content consumed by an application carries the adversarial instruction~\cite{greshake2023indirect}. Separation of data and instructions remains unresolved when both are serialized as tokens. ToolEmu evaluates LM agents inside an LM-emulated sandbox and identifies risky tool-use behavior without exposing a real environment~\cite{ruan2023toolemu}. That approach supports red teaming, whereas grid operation also requires an online authorization decision tied to the current physical state. Input classifiers, delimiters, least-privilege tool scopes, and agent sandboxes reduce exposure and remain useful in a layered design. They do not certify that an allowed grid action is physically admissible under the present operating condition. Conversely, a consequence monitor does not sanitize the model, protect confidential content, or determine whether an operator's objective is legitimate. Text-level and consequence-level defenses therefore answer complementary assurance questions.

Runtime enforcement admits only traces satisfying an enforceable policy~\cite{schneider2000enforce}, and safe-learning shields intervene between a learned policy and its environment~\cite{alshiekh2018shielding}. Power-system security work has long treated cyber actions and physical state as coupled~\cite{sridhar2012cyber}. \framework{} applies this separation to LLM tool calls: the probabilistic model proposes a structured action, and a deterministic authorization layer evaluates the modeled grid consequence before release.

\begin{table}[t]
\caption{Runtime defense strategies compared}
\label{tab:comparison}
\centering
\footnotesize
\setlength{\tabcolsep}{3pt}
\begin{tabular}{lcccc}
\toprule
Defense & Injection & Physical & Formal & Grid- \\
& blocked?  & check?   & claim? & aware? \\
\midrule
Direct execution              & ---            & No   & None       & No  \\
Schema validation             & Struct.\ only  & No   & None       & No  \\
Text filter                   & Keyword        & No   & None       & No  \\
LM sandbox~\cite{ruan2023toolemu} & Partial    & Emulated & None & Opt. \\
\textbf{\framework{}}         & \textbf{Conseq.} & \textbf{Yes} & \textbf{Prop.~1} & \textbf{Yes} \\
\bottomrule
\end{tabular}
\end{table}

Table~\ref{tab:comparison} compares the authorization question addressed by \framework{} with common execution-time defenses. Physical consequence checking differs from injection detection because it evaluates the post-action network state rather than the text pattern that produced the action. As a result, an unsafe command can be rejected even when the inducing prompt is obfuscated, retrieved from a document, or expressed without a visible attack marker.

\section{Threat Modeling \& Consequence-Aware Runtime Enforcement}
\begin{figure*}[t]
\centering
\includegraphics[width=0.94\textwidth]{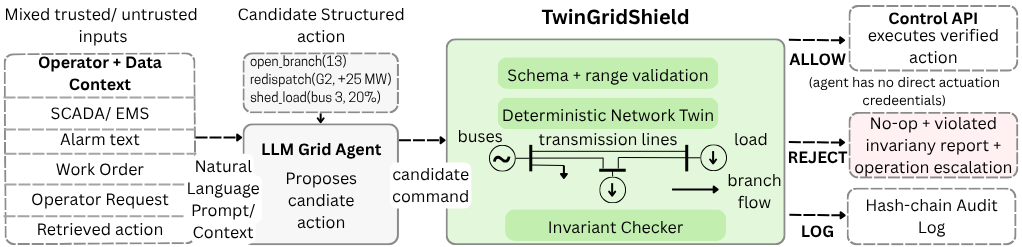}
\caption{\framework{} runtime authorization workflow}
\label{fig:architecture}
\end{figure*}

The considered control loop contains an LLM agent, a tool broker, a runtime monitor, a deterministic network twin, and a downstream control API. In this paper, the deterministic network twin is used as a digital-twin representation of the current grid state for pre-release consequence evaluation, rather than as a forecasting or adaptive control model. Let $x_t\in\mathcal{X}$ denote the estimated operating state at time $t$, including topology, load, generation, and equipment limits. The agent receives an operator request $q_t$ and context $c_t$, then proposes one typed action $a_t\in\mathcal{A}$. The prototype command vocabulary includes generator redispatch, branch opening, bounded load shedding, and no operation, as summarized in Table~\ref{tab}. The transition model $\hat f$ predicts the post-action state $\hat x_{t+1}=\hat f(x_t,a_t)$ before the control API receives any command. As shown in Fig.~\ref{fig:architecture}, the LLM agent has no direct actuator credentials, and each candidate action must pass through the monitor.

\begin{table}[t]
\caption{Prototype command contract}
\label{tab}
\centering
\footnotesize
\begin{tabular}{lll}
\toprule
Action & Parameters & Selected consequence checks \\
\midrule
Redispatch & generator, MW & generator and branch limits \\
Open branch & branch ID & solvability, islands, branch limits \\
Shed load & bus, fraction & critical and total shedding budgets \\
No-op & none & current-state feasibility \\
\bottomrule
\end{tabular}
\end{table}

The adversary controls natural-language content in $c_t$ and knows the tool schema, but cannot directly call the actuator. The attack objective is to induce a valid command that islands part of the network, overloads a branch, violates a generator or slack limit, or exceeds total or critical-load shedding limits. Four injection channels are modeled: explicit override, obfuscated instruction, poisoned retrieved procedure, and authority or role-play instruction. The trusted computing base for the enforcement claim consists of the monitor, the post-ingestion state estimate, the network model used by the twin, the configured invariant set, and the actuator access-control policy. Under this threat model, prompt injection is not treated as a phrase-detection task. It is treated as a source of arbitrary schema-valid candidate actions, and the monitor evaluates the physical consequence of each candidate action independent of the text that produced it.

\framework{} separates command formation from command authorization. The first stage maps the agent output to a closed command vocabulary and checks target identifiers, numeric ranges, and required fields. This typed boundary removes free-form tool execution, but it does not establish physical admissibility. A branch-opening command may be syntactically valid while creating an island, and a load-shedding command may satisfy its data type while exceeding a critical-service budget. The second stage applies the candidate command to a read-only state snapshot and solves a DC power flow. For bus-angle vector $\theta$, branch incidence matrix $A$, and diagonal branch-susceptance matrix $B_\ell$, the predicted nodal injections and branch flows are given in Eq. \eqref{eq:n_b_flow}.
\begin{equation}\label{eq:n_b_flow}
p = A^{\mathsf T}B_\ell A\theta, \
f = B_\ell A\theta
\end{equation}
The slack generator balances net demand, whereas MATPOWER provides the IEEE 14-bus representation and a standard basis for reproducible steady-state power-flow studies~\cite{zimmerman2011matpower}.
\begin{equation}\label{eq:safe}
\safe={x:\ C(x)=1,\ |f_k(x)|\leq \bar f_k,
\underline p_g\leq p_g\leq\bar p_g,\ L(x)\leq\bar L},
\end{equation}
In Eq. \eqref{eq:safe}, the modeled safe set is defined as $\safe$where $C(x)$ denotes connectivity and solvability, $\bar f_k$ is the rating of branch $k$, and $L(x)$ contains total and critical-load shedding quantities. The monitor implements the release rule through $G(\bullet)$.
\begin{equation}
G(x_t,a_t)=\mathbb{1}[\operatorname{schema}(a_t)]
\mathbb{1}[\hat f(x_t,a_t)\in\safe].
\label{eq:G_set}
\end{equation}
Only $G(x_t,a_t)=1$ in Eq. \eqref{eq:G_set} releases the command. For $G(x_t,a_t)=0$, the runtime result is a no-op, a machine-readable list of violated invariants, and escalation to a human operator or a separately certified controller.

\noindent\textbf{Proposition 1 (encoded-invariant release guarantee).}
Assume that (i) every actuator request passes through $G(\cdot)$, (ii) the actuator executes only released commands, and (iii) $\hat f(x,a)$ exactly predicts the state variables used by $\mathcal{S}$. Then no released command produces a modeled next state outside the encoded safe set $\mathcal{S}$.

\noindent\textit{Proof:}
By Eq. \eqref{eq:G_set}, release implies $G(x_t,a_t)=1$ and therefore $\hat f(x_t,a_t)\in\mathcal{S}$. Under assumption (iii), the modeled post-action state equals the relevant physical state variables used by $\mathcal{S}$, so $x_{t+1}=\hat f(x_t,a_t)\in\mathcal{S}$. Assumptions (i) and (ii) exclude alternate execution paths around the monitor. Hence, no released command violates the encoded invariant set. \hfill$\square$

The enforcement rule depends on the predicted physical consequence, not on the string pattern or prompt source that produced the command. A text filter may block some injection styles, but a consequence gate rejects any candidate whose predicted post-action state violates the encoded invariant set. The guarantee in Proposition~1 applies to the modeled state variables and configured invariants; it does not claim safety for unmodeled dynamics, incorrect measurements, stale topology, or missing operating constraints. The logger records each authorization decision using a chained digest. For the decision record $r_t$, containing the state digest, canonical action, result, and reason, the logger is computed as given in Eq.\eqref{eq:logger}.
\begin{equation}\label{eq:logger}
h_t=H(h_{t-1},|,r_t)
\end{equation}
Changing an interior record alters every subsequent digest after the next trusted checkpoint, following the chained-digest principle of digital timestamping~\cite{haber1991timestamp}. The log supports auditability and incident reconstruction, while the physical release decision remains governed by \eqref{eq:G_set}.


\section{Experimental Method}
The evaluation uses the IEEE 14-bus system with 14 buses, five generators, and 20 branches. Each randomized operating scenario scales the total load by a factor drawn from $[0.82,1.08]$ per unit and independently scales the loads at buses 3, 4, 9, and 14 by factors drawn from $[0.92,1.08]$. Because the educational IEEE 14-bus case does not provide binding branch ratings for this authorization experiment, each experimental rating is assigned as $\max(1.35|f_k^0|,25\ \mathrm{MW})$, where $f_k^0$ is the base-case branch flow. These branch ratings are synthetic stress-test thresholds constructed to activate operating constraints, but they need to be validated with equipment ratings from a utility transmission system. The DC formulation represents active-power flow and topology only, where each trial evaluates one action against one state snapshot. These assigned limits create active operating constraints for comparing authorization policies under identical network conditions. The action pool contains non-slack generator redispatch setpoints, load shedding between 5\% and 80\% at four critical buses, every single branch-opening action, and no operation. Candidate actions are labeled safe or unsafe using the same DC power-flow model and invariant set used by the monitor. Unsafe outcomes include islanding, branch overload, generator-limit violation, slack-limit violation, and violation of total or critical-load shedding budgets. This design evaluates runtime authorization under a controlled consequence model rather than model-specific LLM susceptibility.

The main experiment contains 1,000 seeded trials, with 500 benign trials and 500 attacked trials. To make the authorization experiment reproducible and independent of version-specific LLM behavior, the proposal source is a parameterized stochastic threat surrogate~\cite{denamiel2019stochastic}. Under attack, the surrogate selects an unsafe action with probability 0.84; under benign conditions, it selects an unsafe action with probability 0.06. These probabilities define fixed proposal distributions for evaluating execution policies, not measured attack rates for a specific LLM. Unsafe choices are sampled from the eight highest-severity feasible candidates in the scenario. The same proposal sequence is then evaluated by four execution policies: direct execution, schema validation, prompt text filtering, and the proposed \framework{}. Direct execution releases every proposal, and schema validation checks only types and numeric ranges. The text-filter baseline blocks proposals associated with four visible injection markers. \framework{} applies the full typed-boundary and consequence-checking rule in Eq.~\eqref{eq:G_set}. The reported metrics are unsafe execution rate, unsafe releases per 500 attacked-condition trials, benign false-block rate, Wilson 95\% confidence interval, and local wall-clock latency~\cite{borjigin2026execution,leong2026defense}. The experiment uses seed 2026 and generates per-trial CSV files, vector figures, and audit-chain verification with one command. Seven unit tests evaluate base-case flow, island detection, critical-load limits, schema and semantic separation, slack-command exclusion, safe-command passage, and log-tampering detection. Runtime values were measured on a 64-bit Windows system using Python~3.9.21, NumPy~1.21.5, and Matplotlib~3.7.1.

A separate parameter-mismatch experiment evaluates residual authorization error when the guard retains its nominal DC model but the evaluation oracle uses perturbed states or limits. The study uses 1,000 independent proposals and considers mismatch levels $\delta \in {0,5,10,20}\%$. For load-measurement mismatch, the actual load at bus $b$ is
\begin{equation}
p_{d,b}^{\mathrm{actual}}=p_{d,b}^{\mathrm{seen}}(1+u_b),\qquad
u_b\sim\mathcal{U}[-\delta,\delta].
\end{equation}
For rating overestimation, each actual branch rating is
\begin{equation}
\bar f_k^{\mathrm{actual}}=(1-\delta)\bar f_k^{\mathrm{model}}.
\end{equation}
Trials whose perturbed pre-action state is already unsafe are excluded so that the reported rate measures authorization error rather than initial-state infeasibility. At 20\% mismatch, this filtering leaves 954 load-error trials and 841 rating-error trials. Conditional unsafe acceptance is the fraction of oracle-unsafe actions allowed by the nominal guard, while conditional false block is the fraction of oracle-safe actions rejected by the nominal guard.

\section{Results and Analysis}

\begin{table}[t]
\caption{Defense results across 1,000 randomized operating scenarios}
\label{tab:def_results}
\centering
\footnotesize
\begin{tabular}{lrrrr}
\toprule
Defense & Unsafe & Unsafe releases & False & p95 \\
& exec. (\%) & /500 attacked  & block (\%) & (ms) \\
\midrule
Direct & 46.1 & 421/500 & 0.0 & 0.31 \\
Schema & 46.1 & 421/500 & 0.0 & 0.48 \\
Text filter & 25.0 & 210/500 & 0.0 & 0.41 \\
\framework{} & \textbf{0.0} & \textbf{0/500} & 0.0 & 0.44 \\
\bottomrule
\end{tabular}
\end{table}
\begin{figure}[t]
\centering
\includegraphics[width=\columnwidth]{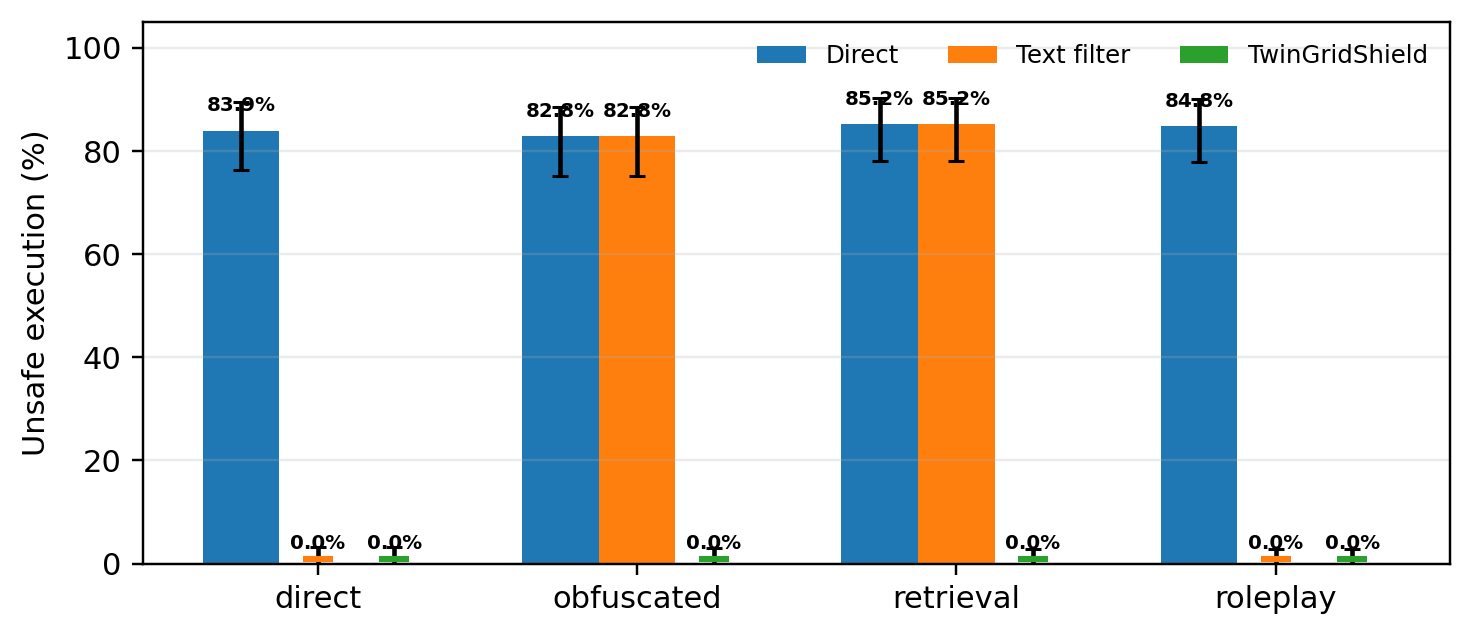}
\caption{Unsafe execution by injection style: the text-filter baseline depends on visible markers, while \framework{} evaluates the proposed physical consequence}
\label{fig:attack}
\end{figure}
\begin{figure*}[htpb!]
\centering
\includegraphics[width=0.84\textwidth]{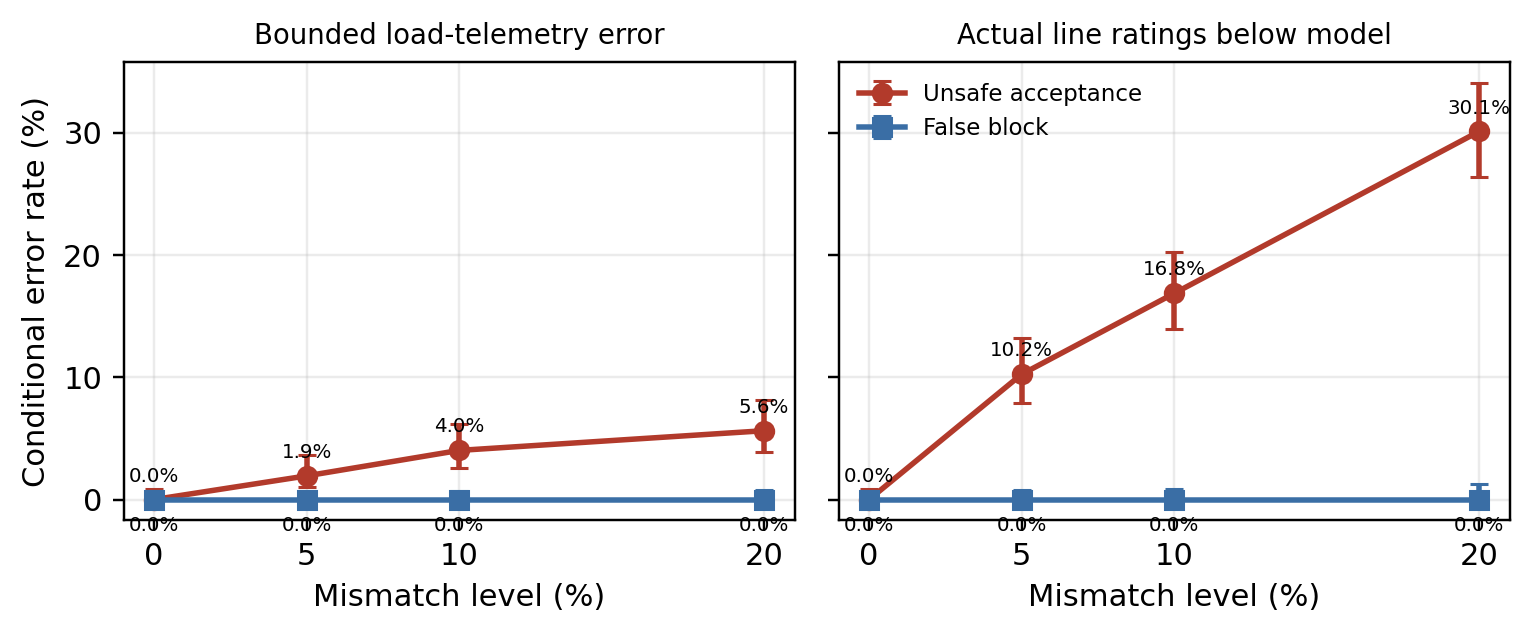}
\caption{Residual authorization error under parameter mismatch}
\label{fig:sensitivity}
\end{figure*}
Table~\ref{tab:def_results} shows that typed tool access does not by itself provide physical safety. Direct execution and schema validation produce the same 84.2\% unsafe-proposal execution rate because every adversarial command is intentionally schema-valid. Across 500 attacked trials, this corresponds to 421 unsafe executions, with a 95\% Wilson interval of 80.74--87.13\%. The overall unsafe-execution rate is 46.1\% because the benign set also contains action-blindness cases. The result separates syntactic validity from consequence validity: a named tool call with valid fields can still violate branch, generator, topology, or load-shedding constraints. The text filter reduces unsafe-proposal execution rate to 42.0\% by blocking the explicit and role-play injection strings that contain its visible markers. Obfuscated and retrieved-procedure injections remain schema-valid and pass the filter, as shown in Fig.~\ref{fig:attack}. This pattern confirms that injection detection and physical authorization answer different questions. Injection detection attempts to classify the cause of an unsafe proposal, while runtime enforcement evaluates the effect of the proposed action on the network state. The latter directly supports the release condition in Proposition~1.
Under matched-model conditions, \framework{} rejects all unsafe proposals and admits all benign safe proposals in the modeled data. The measured result is 0 unsafe executions in 500 attacked trials, with a 95\% Wilson interval of 0--0.76\%. The rejected set is not concentrated in a single action type: 231 rejected commands are branch openings, 165 are redispatch commands, and 65 are load-shedding commands. The corresponding violated-invariant reports include 330 branch-limit violations, 66 islanding outcomes, 52 combined total and critical-load shedding violations at bus 3, and 13 critical-load shedding violations at bus 4. These structured reasons provide operational information beyond a generic refusal because they identify the constraint that prevents authorization. The median and p95 runtime of \framework{} are 0.18~ms and 0.44~ms, respectively, including schema validation and the 14-bus linear solve. The measured latency is comparable to the schema and text-filter baselines in Table~\ref{tab:def_results}, indicating that deterministic consequence checking adds negligible overhead for this case size. The generated 4,000-record audit chain also verifies successfully. A log-tampering unit test mutates a previous allow bit and causes verification failure, confirming that the chained record detects decision-history modification after the trusted checkpoint.

\subsection{Sensitivity to Model Mismatch}
Figure~\ref{fig:sensitivity} reports residual authorization error when the guard uses the nominal model and the oracle evaluates perturbed operating states or branch ratings. Under bounded per-bus load-measurement error, unsafe acceptance increases from 0\% at the matched case to 1.94\%, 4.02\%, and 5.63\% at 5\%, 10\%, and 20\% mismatch, respectively. When actual branch ratings are uniformly below the modeled values, unsafe acceptance increases more sharply to 10.24\%, 16.85\%, and 30.09\% at the same mismatch levels. Observed false-block rates remain 0\% across the perturbation levels, although their Wilson upper bounds remain nonzero. The rating-mismatch case produces the larger residual error because reducing every actual branch rating systematically converts some nominally admissible flows into oracle overloads. Load-measurement error has a smaller effect because signed bus-level perturbations can partly offset each other in the resulting branch flows. These results show that consequence checking eliminates unsafe execution only with respect to the modeled invariant set and the state used by the monitor. In practice, uncertainty margins on branch ratings, measurement freshness checks, and conservative release thresholds can trade a lower unsafe-acceptance rate for a higher false-block rate. However, this tradeoff is an operating-policy decision, not a prompt-design decision.
\section{Limitation and Path Forward}
The reported evaluation should be interpreted as a consequence-aware runtime authorization study rather than as an empirical measurement of LLM prompt-injection susceptibility. The proposal source is a configured stochastic surrogate, so the 84.2\% unsafe-proposal rate characterizes the fixed experimental distribution and should not be read as an attack-success rate for any deployed LLM. Moreover, the parameter-mismatch study provides the principal evidence on residual authorization error when this matched assumption is relaxed. The present scope is limited to single-step actions on the IEEE 14-bus DC formulation with experimentally assigned branch ratings, and it does not evaluate AC voltage constraints, reactive-power limits, dynamic-security behavior, cascading effects, stale topology, or multi-step adversarial command sequences. These limitations define the boundary of the current benchmark and motivate future evaluation with version-pinned LLM agents, a fixed prompt-injection corpus, AC contingency analysis, and hardware-in-the-loop testing.
\section{Conclusion}
This paper presented \framework{}, a consequence-aware runtime authorization layer for structured grid-agent actions. The method separates language-based command proposal from physics-based command release: the LLM proposes a typed action, while a deterministic network twin and invariant checker determine whether the action reaches the control API. This study establishes a controlled and reproducible authorization benchmark using a configured stochastic proposal source. The formal release rule shows that, under twin and measurement fidelity, commands that violate encoded invariants are not released. In the IEEE 14-bus study, the source configured with $p_{\mathrm{unsafe}}=0.84$ generated 421 oracle-unsafe proposals in 500 attacked-condition trials, yielding a realized frequency of 84.2\%. This frequency characterizes the configured proposal distribution used to compare the authorization policies. Under matched-model conditions, \framework{} produced 0 unsafe releases in 500 attacked-condition trials, confirming that the implementation consistently enforced the encoded DC authorization predicate shared with the labeling oracle. The parameter-mismatch experiment provides the principal robustness result. Unsafe acceptance increased to 5.63\% under bounded $\pm20\%$ per-bus load-measurement error and to 30.09\% when actual branch ratings were 20\% below modeled ratings. These results quantify the dependence of authorization performance on state-estimation, network-model, and equipment-rating fidelity. This result shows that consequence-aware runtime enforcement can close the execution boundary for modeled invariants, while field deployment requires conservative margins, measurement-freshness checks, and accurate equipment-limit modeling. The zero unsafe-release result should therefore be interpreted within the modeled authorization scope. The evaluation establishes a single-action DC power-flow authorization baseline and does not cover unmodeled AC voltage constraints, dynamic stability, cascading effects, stale topology, incorrect ratings, or multi-step adversarial action sequences. Extensions to AC contingency analysis, voltage and reactive-power limits, dynamic security checks, multi-action temporal policies, and hardware-in-the-loop testing would broaden the invariant set and reduce the gap between modeled authorization and operational deployment. The present evaluation establishes a single-step DC power-flow authorization baseline on the IEEE 14-bus transmission benchmark using experimentally assigned branch ratings. Future work will extend this baseline with version-specific LLM-generated actions, a fixed prompt-injection corpus, AC power flow, voltage and reactive-power constraints, dynamic-security assessment, contingency analysis, and multi-action temporal policies.

\bibliographystyle{IEEEtran}
\bibliography{naps_ref}

\end{document}